\documentclass[10pt,twocolumn,letterpaper]{article}

\usepackage[pagenumbers]{cvpr} %

\usepackage{booktabs}
\usepackage{makecell}
\usepackage{multirow}

\definecolor{cvprblue}{rgb}{0.21,0.49,0.74}
\usepackage[pagebackref,breaklinks,colorlinks,allcolors=cvprblue]{hyperref}

\title{GaussianWorld: Gaussian World Model for Streaming 3D Occupancy Prediction}

\author{
Sicheng Zuo\footnotemark[1]\quad Wenzhao Zheng{\footnotemark[1] $^,$\footnotemark[2]}\quad Yuanhui Huang\quad Jie Zhou\quad Jiwen Lu \\
Department of Automation, Tsinghua University, China \\
\texttt{\small zsc23@mails.tsinghua.edu.cn; wenzhao.zheng@outlook.com} \\
}

\begin{document}

\twocolumn[{%
\renewcommand\twocolumn[1][]{#1}%
\vspace{-12mm}
\maketitle
\vspace{-8mm}
\begin{center}
    \centering
    \includegraphics[width=\linewidth]{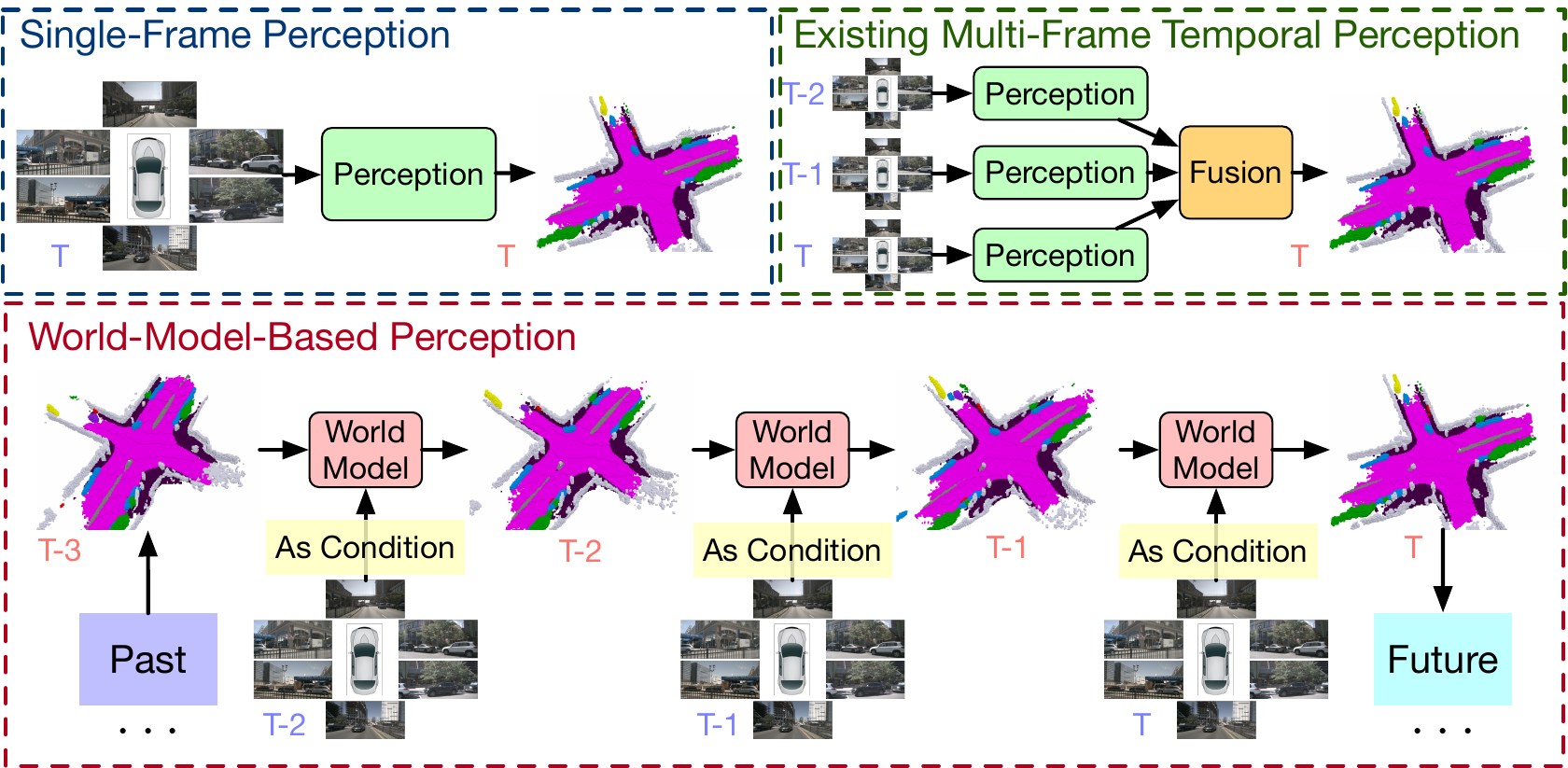}
    \vspace{-5mm}
    \captionof{figure}{
While single-frame 3D occupancy prediction methods demonstrate strong performance~\cite{occformer,surroundocc}, the incorporation of temporal information can further improve the results~\cite{cvt-occ}. 
However, most existing methods fuse past scene representations~\cite{bevformer,fb-occ} to infer the current 3D occupancy, which ignores the continuity of driving scenarios and introduces additional computations.
Differently, we propose a world-model-based framework for streaming 3D occupancy prediction and explicitly model the scene evolutions using the current camera observations as inputs. 
Our framework improves the performance of existing methods without additional computation overhead.
}
\label{teaser}
\end{center}%
}]

\renewcommand{\thefootnote}{\fnsymbol{footnote}}
\footnotetext[1]{Equal contribution. $\dagger$Project leader.}
\renewcommand{\thefootnote}{\arabic{footnote}}

\begin{abstract}
3D occupancy prediction is important for autonomous driving due to its comprehensive perception of the surroundings.
To incorporate sequential inputs, most existing methods fuse representations from previous frames to infer the current 3D occupancy.
However, they fail to consider the continuity of driving scenarios and ignore the strong prior provided by the evolution of 3D scenes (e.g., only dynamic objects move).
In this paper, we propose a world-model-based framework to exploit the scene evolution for perception.
We reformulate 3D occupancy prediction as a 4D occupancy forecasting problem conditioned on the current sensor input.
We decompose the scene evolution into three factors: 1) ego motion alignment of static scenes; 2) local movements of dynamic objects; and 3) completion of newly-observed scenes.
We then employ a Gaussian world model (GaussianWorld) to explicitly exploit these priors and infer the scene evolution in the 3D Gaussian space considering the current RGB observation.
We evaluate the effectiveness of our framework on the widely used nuScenes dataset.
Our GaussianWorld improves the performance of the single-frame counterpart by over 2\% in mIoU without introducing additional computations.
Code: \url{https://github.com/zuosc19/GaussianWorld}.
\end{abstract}

\vspace{-3mm}
\section{Introduction}
\label{sec:intro}

Vision-centric 3D occupancy prediction has recently gained much attention due to its critical applications in autonomous driving~\cite{monoscene,tpvformer,occformer,occworld}.
The task aims to estimate the occupancy status and semantic labels of each voxel in a 3D environment based on visual inputs~\cite{openoccupancy,surroundocc,tian2023occ3d,scene_as_occ}.
3D occupancy provides a more fine-grained semantic and structural description of the scene, which is important for developing safe and robust autonomous driving systems~\cite{hu2023uniAD,vad,occworld}. 

Leveraging temporal inputs is important for 3D occupancy prediction as it provides sufficient historical context for understanding the scene evolution~\cite{bevformer,bevdet4d,sparse4d,streampetr}.
Most existing methods follow a conventional pipeline of perception, transformation, and fusion~\cite{driving-in-occworld,driveworld,fb-occ,cvt-occ}.
Given sequential inputs, the perception module obtains the scene representation of each frame independently, such as the bird's eye view (BEV) features~\cite{li2023bevstereo,bevdet,bevformer} or cost volume features~\cite{voxformer,panoocc,octreeocc}.
The transformation module then aligns the multi-frame features based on the ego trajectory, and the fusion module fuses the aligned representations to infer the current 3D occupancy. 
However, these methods fail to consider the inherent continuity and simplicity of the evolution of driving scenarios.
The driving scene representations in adjacent frames are supposed to be closely correlated to each other, as the scene evolution generally originates only from the movements of the ego vehicle and other dynamic objects.
The direct fusion of multi-frame representations ignores this strong prior provided by the evolution of 3D scenes, i.e. static object coherence and dynamic object motion, which makes it difficult for the model to understand the development of driving scenes.
Furthermore, this design increases the complexity and computational effort of temporal modeling, reducing its efficiency and effectiveness.

In this paper, we explore a world-model-based paradigm,  GaussianWorld,  to exploit the scene evolution for perception.
We adopt explicit 3D Gaussians as the scene representation~\cite{gaussianformer} over the conventional implicit BEV/Voxel representation~\cite{bevformer,bevdet,bevdepth}, which enables the explicit and continuous modeling of object movements. 
To facilitate perception, we reformulate 3D occupancy prediction as a 4D occupancy forecasting problem conditioned on the current sensor input.
Given the historical 3D Gaussians and the current visual input, GaussianWorld aims to predict how the scene evolves and forecast the current occupancy.
To achieve this, we decompose the scene evolution into three factors: 1) ego motion alignment of static scenes; 2) local movements of dynamic objects; and 3) completion of newly-observed areas.
To model these factors in the 3D Gaussian space, we first align historical 3D Gaussians to the current frame based on the ego trajectory.
We also complete the newly-observed areas with random Gaussians, facilitating the perception of these new areas.
We propose a unified refinement layer to simultaneously model the progression of historical Gaussians and the perception of newly-completed Gaussians.
Finally, we utilize all refined Gaussians to predict the scene evolution and determine the current occupancy.

To demonstrate the effectiveness of GaussianWorld, we have conducted extensive experiments on the widely used nuScenes~\cite{nuscenes} dataset.
As shown in Figure~\ref{teaser}, our GaussianWorld can effectively predict the scene evolution and improve the single-frame occupancy prediction by over 2\% in mIoU without introducing additional computations.

\section{Related Work}
\label{sec:formatting}
\textbf{3D Occupancy Prediction}: 3D occupancy prediction has gained increasing attention since it describes the fine-grained 3D structure of driving scenes~\cite{monoscene,tpvformer,occformer,occworld}.
Early methods utilized the LiDAR points as inputs to complete the semantics of the entire scene~\cite{js3c,lmscnet,pointocc}.
Recent works have focused on the more challenging vision-based 3D occupancy prediction~\cite{monoscene,tpvformer}.
A straightforward approach is to obtain dense voxel representations from images to predict 3D occupancy~\cite{voxformer,symphonize,cotr}. 
Considering the sparsity of occupied voxels in driving scenes, other methods have explored more efficient representations, such as Tri-Perspective View (TPV)~\cite{tpvformer}, 3D Gaussians~\cite{gaussianformer}, and points~\cite{occaspoints}.
However, these methods are limited to single-frame perception, neglecting the incorporation of temporal information.
Recently, CVT-Occ~\cite{cvt-occ} explored the temporal fusion of 3D volume features to improve 3D occupancy prediction.
Despite this, the approach fails to consider the close correlation between consecutive frames in driving scenes.
Considering this, we propose a world-model-based framework to exploit the scene evolution for perception.

\textbf{World Models for Autonomous Driving}: A world model is usually defined as a predictive model of the future based on historical observations and actions~\cite{ha2018world}.
Current applications of world models in autonomous driving mainly include driving scene generation~\cite{wang2023drivedreamer,hu2023gaia,gao2023magicdrive,yang2023bevcontrol}, planning~\cite{occworld,mile,drivewm}, and representation learning~\cite{vidar,driveworld}. 
World models based on advanced generative models~\cite{rombach2022high,brown2020language} can generate diverse driving sequences across images~\cite{yang2024genad,gao2024vista}, points~\cite{copilot4d}, and occupancy~\cite{occworld}.
By jointly modeling the scene evolution and ego motion, World models can learn an effective driving policy for planning~\cite{occworld}.
Additionally, world modeling has been employed as a 4D pre-training task to learn a general scene representation~\cite{vidar,driveworld}.
However, how world models can facilitate perception tasks has not been explored. 
In this paper, we employ a Gaussian world model to learn the scene evolution and enhance 4D occupancy forecasting based on current RGB observations.

\textbf{Temporal Modeling for 3D Perception}: Leveraging temporal information is crucial for 3D perception~\cite{sparse4d,streampetr}.
A common approach involves fusing multi-frame scene representations to enhance perception tasks~\cite{bevdet4d,bevformer,fb-occ}.
It aligns multi-frame representations to the current time and aggregates temporal information.
However, this design fails to consider the continuity and simplicity of driving scenarios, which limits the performance of temporal modeling.
StreamPETR~\cite{streampetr} proposed a novel object-centric temporal modeling mechanism for streaming 3D prediction.
As it uses object queries as the scene representation, it can only implicitly model the motion of dynamic objects and is not suitable for dense occupancy prediction.
Differently, we propose a Gaussian world model to predict the scene evolution and forecast 4D occupancy in a streaming manner.

\begin{figure*}[t]
\centering
\includegraphics[width=1.0\textwidth]{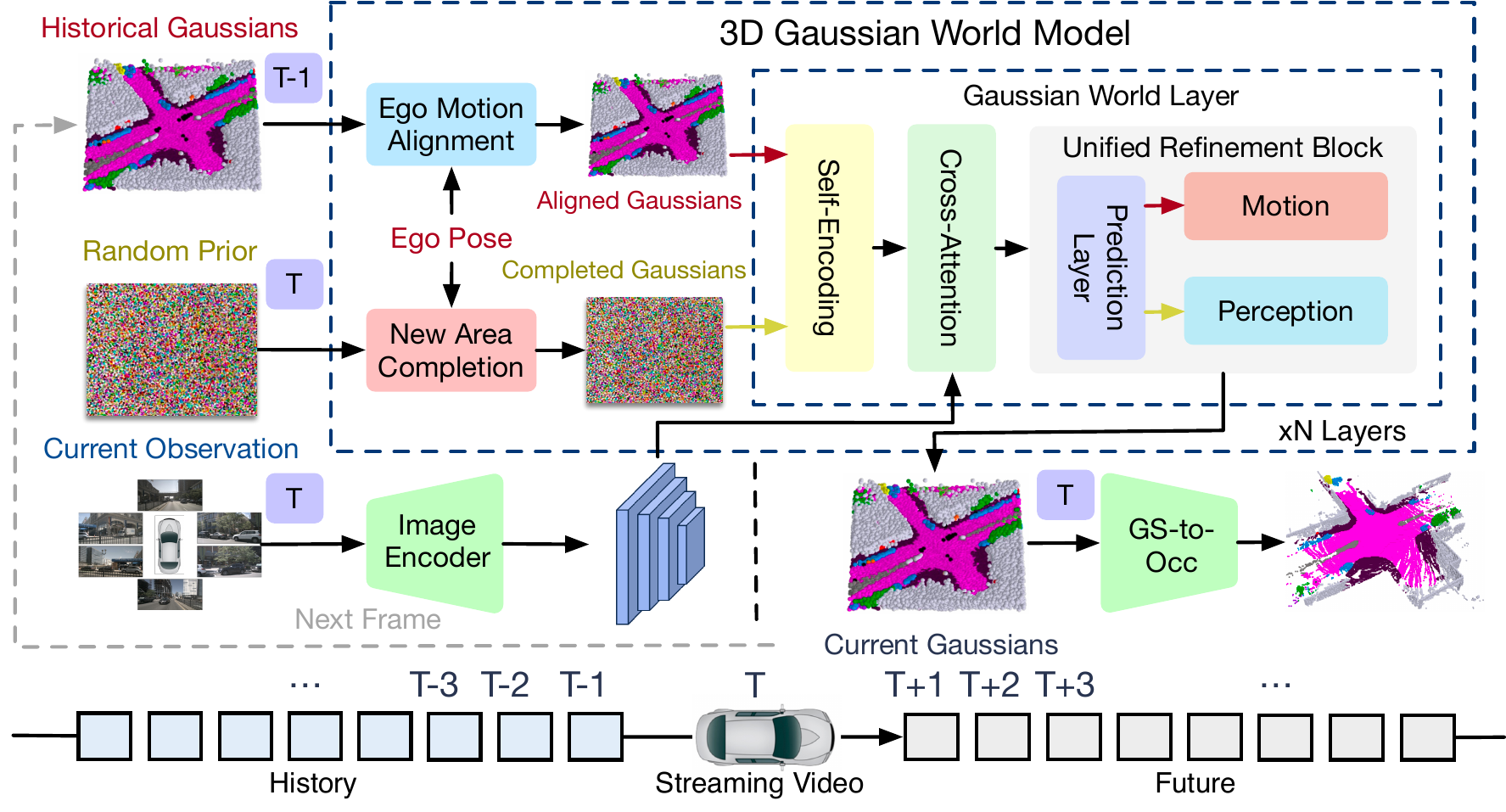}
\vspace{-3mm}
\caption{\textbf{Framework of our GaussianWorld for streaming 3D semantic occupancy prediction.}
As the ego vehicle shifts from the last frame to the current frame, we first align historical Gaussians to the current time and complete newly-observed areas with random Gaussians.
We then utilize several Gaussian world layers composed of self-encoding, cross-attention, and unified refinement blocks to simultaneously predict the development of historical Gaussians and the properties of completed Gaussians.
The refined Gaussians can model the scene evolution and generate the current occupancy.
}
\label{fig:framework}
\vspace{-5mm}
\end{figure*}

\section{Proposed Approach}

\subsection{World Models for Perception}
Precisely perceiving 3D scenes is crucial for developing reliable autonomous driving systems.
It aims to predict the geometry and semantics of 3D scenes to support the subsequent prediction and planning tasks.
The perception model $A$ takes the sensor inputs $\{\mathbf{x}^T, \mathbf{x}^{T-1}, ..., \mathbf{x}^{T-t}\}$ from last $t$ frames and the current frame $T$ to obtain perception $\mathbf{y}^T$ :
\begin{equation}
\label{eq:perceive-temporal}
  \mathbf{y}^{T} = A(\{\mathbf{x}^T, ..., \mathbf{x}^{T-t}\},\  \{\mathbf{p}^T, ..., \mathbf{p}^{T-t}\}),
\end{equation}
where $\mathbf{p}^t$ denotes the ego position of the time stamp $t$.

The conventional pipeline of temporal modeling for perception consists of three stages: perception, transformation, and fusion.
The perception module $P_{er}$ first extracts the scene representation $\mathbf{z}$ for each frame independently.
As the ego vehicle advances, the ego-centric representations across frames are misaligned.
The transformation module $T_{rans}$ addresses this by aligning past features to the current frame based on the ego trajectory.
The fusion module $F_{use}$ then integrates the aligned multi-frame representations for perception.
The conventional pipeline can be formulated as:
\begin{equation}
\begin{aligned}
\label{eq:perceive-temporal-fusion}
  &  \mathbf{z}^n = P_{er}(\mathbf{x}^n), \mathbf{a}^n = T_{rans}(\mathbf{z}^n, \mathbf{p}^n), \\
  &  \mathbf{y}^{T} = F_{use}(\mathbf{a}^T, ..., \mathbf{a}^{T-t}), \\
\end{aligned}
\end{equation}
where $\mathbf{a}^n$ denotes the aligned scene representation of the $n_{th}$ frame, and n ranges from $T-t$ to $T$.

Despite the promising performance of this framework, it fails to consider the inherent continuity and simplicity of driving scenarios.
The evolution of driving scenes typically originates solely from the movement of the ego vehicle and other dynamic objects.
The driving scene representations in adjacent frames are inherently correlated, containing the evolution dynamics and physical laws of the world.
However, directly fusing multi-frame representations ignores this strong prior, thus limiting its performance.

Motivated by this, we explore a world-model-based paradigm to exploit the scene evolution for perception.
The world model enhances perception by learning a straightforward but effective prior for temporal modeling.
We use a percetpion world model $\mathbf{w}$ which predicts the current representation $\mathbf{z}^{T}$ based on previous representations $\mathbf{z}^{T-1}$ and the current sensor input $\mathbf{x}^{T}$:
\begin{equation}
\label{eq:world-model}
  \mathbf{z}^{T} = \mathbf{w}(\mathbf{z}^{T-1}, \mathbf{x}^{T}).
\end{equation}
We further reformulate the 3D perception task as a 4D forecasting problem conditioned on the current sensor input:
\begin{equation}
\label{eq:4d-perception}
  \mathbf{y}^{T} = \mathbf{A}(\mathbf{z}^{T-1}, \mathbf{x}^{T}) = \mathbf{h}(\mathbf{w}(\mathbf{z}^{T-1}, \mathbf{x}^{T})),
\end{equation}
where $\mathbf{h}$ is the perception head based on representation $\mathbf{z}$.

Having obtained the predicted scene representation $\mathbf{z}^{T}$ and the next observation $\mathbf{x}^{T+1}$, we can input them into the world model to predict the next representation $\mathbf{z}^{T+1}$ in a streaming manner.
The world model learns the joint distribution of the scene representation conditioned on the scene evolution and current observations for perception.

\subsection{Explicit Scene Evolution Modeling}
\begin{figure}[t]
\centering
\includegraphics[width=0.475\textwidth]{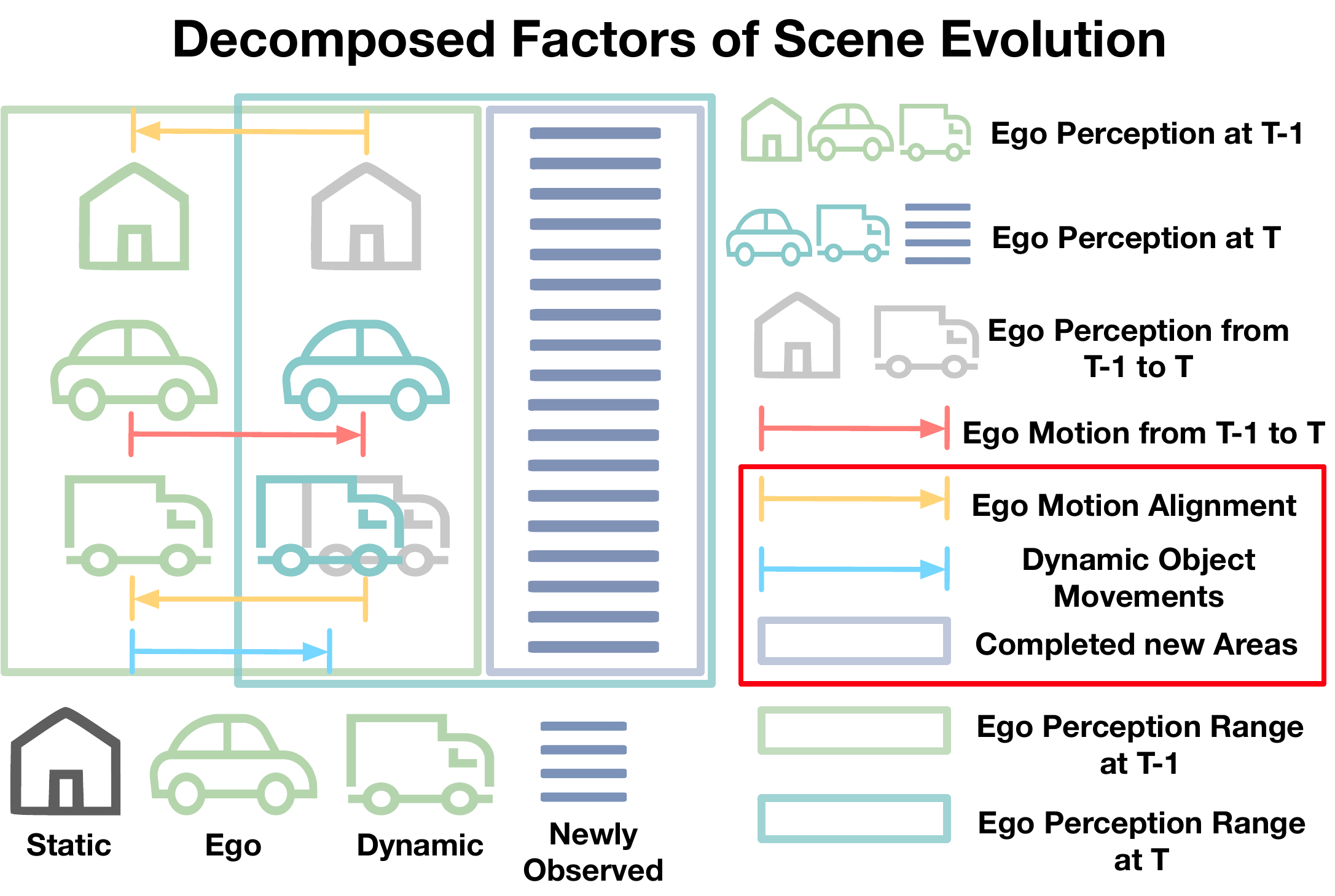}
\vspace{-7mm}
\caption{\textbf{Illustration of the three decomposed factors of scene evolution.}
We decompose the scene evolution into three key factors: {ego-motion alignment of static scenes}, {local movements of dynamic objects}, and {the completion of newly-observed areas}.
}
\label{fig:factors}
\vspace{-6mm}
\end{figure}

The evolution of driving scenes is generally simple and continuous, primarily caused by the movements of dynamic objects.
When adopting an ego-centric scene representation within a certain range, the scene evolution can generally be decomposed into three key factors: 1) \textit{the ego motion alignment of static scenes}; 2) \textit{the local movements of dynamic objects}; and 3) \textit{the completion of newly-observed areas}, as shown in Figure~\ref{fig:factors}.
By modeling these factors, the world model can learn to effectively evolve scenes.

Considering the above decomposition of the scene evolution, we adopt 3D Gaussians as the scene representation\cite{gaussianformer} to explicitly and continuously model the scene evolution.
We describe 3D scenes with a set of sparse 3D semantic Gaussians, where each Gaussian represents a flexible region with explicit position, scale, rotation, and semantic probability.
To learn the scene evolution, we introduce an additional temporal feature attribute to capture the historical information of 3D Gaussians.
The 3D Gaussian representation can be formulated as:
\begin{equation}
\label{eq:gaussian-representation}
  \mathbf{g} = \{\mathbf{p}, \mathbf{s}, \mathbf{r}, \mathbf{c}, \mathbf{f}\},
\end{equation}
where $\mathbf{p}$, $\mathbf{s}$, $\mathbf{r}$, $\mathbf{c}$, $\mathbf{f}$ correspond to the position, scale, rotation, semantic probability, and temporal feature of the 3D Gaussian $\mathbf{g}$, respectively.
We further propose a 3D Gaussian world model, GaussianWorld, to exploit the scene evolution for perception.
The proposed GaussianWorld $\mathbf{w}$ operates on the previous 3D Gaussians $\mathbf{g}^{T-1}$ and the current sensor input $\mathbf{x}^{T}$ to predict the current 3D Gaussians $\mathbf{g}^{T}$:
\begin{equation}
\label{eq:gaussian-world-model}
  \mathbf{g}^{T} = \mathbf{w}(\mathbf{g}^{T-1}, \mathbf{x}^{T}).
\end{equation}
Next, we will introduce how GaussianWorld models the aforementioned decomposed factors of the scene evolution within the 3D Gaussian space.

\textbf{Ego Motion Alignment of Static Scenes}.
The objective of GaussianWorld is to predict the current $\mathbf{g}^{T}$ based on the previous $\mathbf{g}^{T-1}$.
The 3D Gaussian representation $\mathbf{g}$ for each frame represents the scene within a certain range centered on the ego position of the corresponding frame and moving forward results in a global displacement of objects.
GaussianWorld addresses this by employing an alignment module $A_{lign}$ to align the positions of 3D Gaussians $\mathbf{g}^{T-1}$ from the last frame to the current frame.
To achieve this, it performs a global affine transformation to 3D Gaussians of the entire scene based on the ego trajectory.
Formally, given the last frame 3D Gaussians $\mathbf{g}^{T-1}$ and the affine transformation matrix $\mathbf{M}_{ego}$, the aligned 3D Gaussians $\mathbf{g}^{T}_{A}$ can be formulated as:
\begin{equation}
\begin{aligned}
\label{eq:align-module}
  \mathbf{g}^{T}_{A} &= A_{lign}(\mathbf{g}^{T-1}, \mathbf{M}_{ego}), \\
                     &= R_{ef}(\mathbf{g}^{T-1}; \mathbf{M}_{ego}\cdot A_{ttr}(\mathbf{g}^{T-1}; \mathbf{p}); \mathbf{p}),
\end{aligned}
\end{equation}
where $A_{ttr}(\mathbf{g}; \mathbf{p})$ denotes the attribute $\mathbf{p}$ of the 3D Gaussian $\mathbf{g}$, and $R_{ef}(\mathbf{g};\mathbf{n};\mathbf{p})$ denotes updating the attribute $\mathbf{p}$ of the 3D Gaussian $\mathbf{g}$ with $\mathbf{n}$.

\textbf{Local Movements of Dynamic Objects}.
We also consider the local movements of dynamic objects as the scene evolves.
GaussianWorld achieves this by updating the positions of dynamic Gaussians.
The aligned 3D Gaussians $\mathbf{g}^{T}_{A}$ are divided into two mutually exclusive sets based on their semantic probabilities: dynamic Gaussian set $\{\mathbf{g}_D\}$ and static Gaussian set $\{\mathbf{g}_S\}$.
GaussianWorld then employs a motion layer $M_{ove}$ to learn the joint distribution of the aligned 3D Gaussians $\mathbf{g}^{T}_{A}$ and the current observation $\mathbf{x}^{T}$ for predicting the movements of dynamic Gaussians $\{\mathbf{g}_D\}$:
\begin{equation}
\begin{aligned}
\label{eq:move-layer}
  \mathbf{g}^{T}_{M} &= M_{ove}(\mathbf{g}^{T}_{A}, \mathbf{x}_{T}), \\
                     &= R_{ef}(\mathbf{g}^{T}_{A}; E_{nc}(\mathbf{g}^{T}_{A}, \mathbf{x}_{T})\cdot I(\mathbf{g}^T_A\in \{\mathbf{g}_D\}); \mathbf{p}),
\end{aligned}
\end{equation}
where $E_{nc}$ denotes the encoder module that facilitates the high-order interaction between 3D Gaussians and the sensor input, and $I(\cdot)$ is the indicator function. 

\textbf{Completion of Newly-Observed Areas}.
As the ego vehicle shifts to a new position, certain existing areas fall outside the boundary, while some new areas become observable.
We discard the Gaussians that fall outside the boundary and complete the newly-observed areas with randomly initialized Gaussians.
To maintain a consistent number of 3D Gaussians, we uniformly sample an equivalent number of 3D Gaussians $\mathbf{g}^{T}_{I}$ in the newly observed areas.
Subsequently, GaussianWorld employs a perception layer $P_{er}$ to predict all attributes of completed 3D Gaussians $\mathbf{g}^{T}_{I}$ in the newly-observed areas based on the current observation $\mathbf{x}^{T}$:
\begin{equation}
\begin{aligned}
\label{eq:perception-layer}
  \mathbf{g}^{T}_{C} &= P_{er}(\mathbf{g}^{T}_{I}, \mathbf{x}_{T}), \\
                     &= R_{ef}(\mathbf{g}^{T}_{I}; E_{nc}(\mathbf{g}^{T}_{I}, \mathbf{x}_{T}); \{\mathbf{p}, \mathbf{s}, \mathbf{r}, \mathbf{c}, \mathbf{f}\}),
\end{aligned}
\end{equation}
where $E_{nc}$ is the same as the one in $M_{ove}$.

\begin{figure}[t]
\centering
\includegraphics[width=0.475\textwidth]{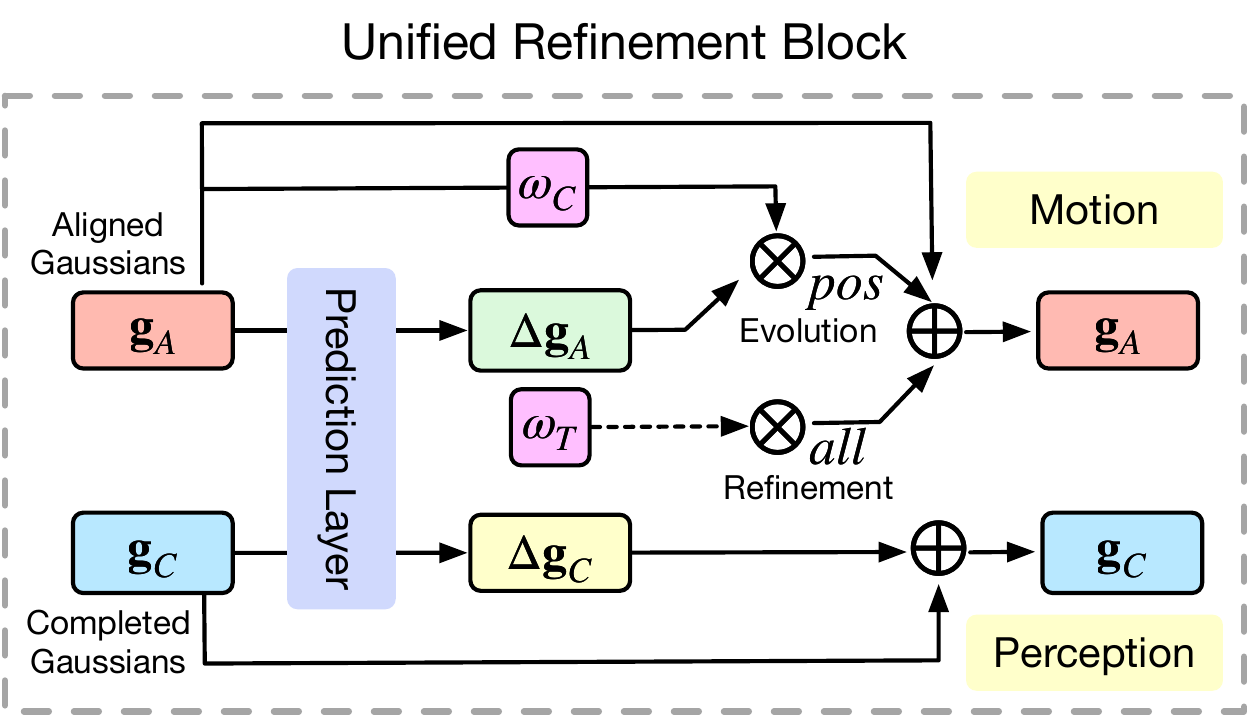}
\vspace{-7mm}
\caption{\textbf{Illustration of the proposed unified refinement block.}
We employ the perception mode to update all attributes of newly completed Gaussians.
We employ the motion mode to predict the evolution of historical Gaussians, where only positions of dynamic Gaussians are updated in the evolution layer $E_{vol}$ and all attributes of historical Gaussians are updated in the refinement layer $R_{efine}$.
}
\label{fig:network}
\vspace{-6mm}
\end{figure}

\definecolor{nbarrier}{RGB}{255, 120, 50}
\definecolor{nbicycle}{RGB}{255, 192, 203}
\definecolor{nbus}{RGB}{255, 255, 0}
\definecolor{ncar}{RGB}{0, 150, 245}
\definecolor{nconstruct}{RGB}{0, 255, 255}
\definecolor{nmotor}{RGB}{200, 180, 0}
\definecolor{npedestrian}{RGB}{255, 0, 0}
\definecolor{ntraffic}{RGB}{255, 240, 150}
\definecolor{ntrailer}{RGB}{135, 60, 0}
\definecolor{ntruck}{RGB}{160, 32, 240}
\definecolor{ndriveable}{RGB}{255, 0, 255}
\definecolor{nother}{RGB}{139, 137, 137}
\definecolor{nsidewalk}{RGB}{75, 0, 75}
\definecolor{nterrain}{RGB}{150, 240, 80}
\definecolor{nmanmade}{RGB}{213, 213, 213}
\definecolor{nvegetation}{RGB}{0, 175, 0}

\newcommand\crule[3][black]{\textcolor{#1}{\rule{#2}{#3}}}
\definecolor{nvcolor}{RGB}{119,185,0}
\definecolor{roadcolor}{RGB}{234,51,246}
\definecolor{sidewalkcolor}{RGB}{68,8,72}
\definecolor{parkingcolor}{RGB}{241,156,249}
\definecolor{othergroundcolor}{RGB}{160,32,76}
\definecolor{buildingcolor}{RGB}{246,202,69}
\definecolor{carcolor}{RGB}{111,149,238}
\definecolor{truckcolor}{RGB}{74,32,172}
\definecolor{bicyclecolor}{RGB}{136,227,242}
\definecolor{motorcyclecolor}{RGB}{37,59,146}
\definecolor{othervehiclecolor}{RGB}{96,81,242}
\definecolor{vegetationcolor}{RGB}{79, 173, 50}
\definecolor{trunkcolor}{RGB}{126, 65, 22}
\definecolor{terraincolor}{RGB}{171, 238, 105}
\definecolor{personcolor}{RGB}{234, 60, 49}
\definecolor{bicyclistcolor}{RGB}{234, 66, 195}
\definecolor{motorcyclistcolor}{RGB}{138, 42, 90}
\definecolor{fencecolor}{RGB}{238, 128, 69}
\definecolor{polecolor}{RGB}{252, 241, 161}
\definecolor{trafficsigncolor}{RGB}{233, 51, 35}
\definecolor{other-struct.color}{RGB}{255, 150, 0}
\definecolor{other-objectcolor}{RGB}{50, 255, 255}
\definecolor{lane-markingcolor}{RGB}{150, 255, 170}
\definecolor{color1}{RGB}{176, 36, 24}
\definecolor{color2}{RGB}{0, 176, 80}
\definecolor{color3}{RGB}{0, 0, 200}
\newcommand{\tbr}[1]{\textbf{\textcolor{color1}{#1}}}
\newcommand{\tbg}[1]{\textbf{\textcolor{color2}{#1}}}
\newcommand{\tbb}[1]{\textbf{\textcolor{color3}{#1}}}

\begin{table*}[t] %
    \caption{\textbf{3D semantic occupancy prediction results on nuScenes validation set.} The original TPVFormer~\cite{tpvformer} is trained with sparse LiDAR segmentation labels, and TPVFormer* is supervised by dense occupancy labels. \textbf{GaussianFormer-B}, \textbf{GaussianFormer-T} denotes the single-frame and temporal fusion variant of GaussianFormer~\cite{gaussianformer}.}
    \setlength{\tabcolsep}{0.003\linewidth}  
    \vspace{-3mm}  
    \renewcommand\arraystretch{1.2}
    \centering
    \resizebox{\textwidth}{!}{
    \begin{tabular}{l|c c | c c c c c c c c c c c c c c c c}
        \toprule
        Method
        &  \makecell{IoU} & \makecell{mIoU}
        & \rotatebox{90}{\textcolor{nbarrier}{$\blacksquare$} barrier}
        & \rotatebox{90}{\textcolor{nbicycle}{$\blacksquare$} bicycle}
        & \rotatebox{90}{\textcolor{nbus}{$\blacksquare$} bus}
        & \rotatebox{90}{\textcolor{ncar}{$\blacksquare$} car}
        & \rotatebox{90}{\textcolor{nconstruct}{$\blacksquare$} const. veh.}
        & \rotatebox{90}{\textcolor{nmotor}{$\blacksquare$} motorcycle}
        & \rotatebox{90}{\textcolor{npedestrian}{$\blacksquare$} pedestrian}
        & \rotatebox{90}{\textcolor{ntraffic}{$\blacksquare$} traffic cone}
        & \rotatebox{90}{\textcolor{ntrailer}{$\blacksquare$} trailer}
        & \rotatebox{90}{\textcolor{ntruck}{$\blacksquare$} truck}
        & \rotatebox{90}{\textcolor{ndriveable}{$\blacksquare$} drive. suf.}
        & \rotatebox{90}{\textcolor{nother}{$\blacksquare$} other flat}
        & \rotatebox{90}{\textcolor{nsidewalk}{$\blacksquare$} sidewalk}
        & \rotatebox{90}{\textcolor{nterrain}{$\blacksquare$} terrain}
        & \rotatebox{90}{\textcolor{nmanmade}{$\blacksquare$} manmade}
        & \rotatebox{90}{\textcolor{nvegetation}{$\blacksquare$} vegetation}
        \\
        \midrule
        MonoScene~\cite{monoscene}  & 23.96 &  7.31 &  4.03 &  0.35 &  8.00 &  8.04 &  2.90 &  0.28 &  1.16 &  0.67 &  4.01 &  4.35 & 27.72 &  5.20 & 15.13 & 11.29 &  9.03 & 14.86 \\
        
        Atlas~\cite{atlas}          & 28.66 & 15.00 & 10.64 &  5.68 & 19.66 & 24.94 &  8.90 &  8.84 &  6.47 &  3.28 & 10.42 & 16.21 & 34.86 & 15.46 & 21.89 & 20.95 & 11.21 & 20.54 \\
        
        BEVFormer~\cite{bevformer}  & 30.50 & 16.75 & 14.22 &  6.58 & 23.46 & 28.28 &  8.66 & 10.77 &  6.64 &  4.05 & 11.20 & 17.78 & 37.28 & 18.00 & 22.88 & 22.17 & \tbb{13.80} & \tbg{22.21}\\
        
        TPVFormer~\cite{tpvformer}  & 11.51 & 11.66 & 16.14 &  7.17 & 22.63	& 17.13 &  8.83 & 11.39 & 10.46 &  8.23 &  9.43 & 17.02 &  8.07 & 13.64 & 13.85 & 10.34 &  4.90 &  7.37\\
        
        TPVFormer*~\cite{tpvformer} & 30.86 & 17.10 & 15.96 &  5.31 & 23.86	& 27.32 &  9.79 &  8.74 &  7.09 &  5.20 & 10.97 & 19.22 & 38.87 & 21.25 & 24.26 & 23.15 & 11.73 & 20.81\\

        OccFormer~\cite{occformer}  & \tbb{31.39} & 19.03 & 18.65 & 10.41 & 23.92 & 30.29 & 10.31 & 14.19 & \tbb{13.59} & 10.13 & 12.49 & 20.77 & 38.78 & 19.79 & 24.19 & 22.21 & 13.48 & 21.35\\

        GaussianFormer~\cite{gaussianformer}
                                    & 29.83 & 19.10 & 19.52 & 11.26 & 26.11 & 29.78 & 10.47 & 13.83 & 12.58 & 8.67 & 12.74 & 21.57 & \tbb{39.63} & 23.28 & 24.46 & 22.99 & 9.59 & 19.12 \\
        
        SurroundOcc~\cite{surroundocc} 
                                      & \tbg{31.49} & \tbb{20.30} & \tbb{20.59} & 11.68      & \tbr{28.06} & \tbb{30.86} & \tbg{10.70}  & 15.14      & \tbr{14.09} & \tbr{12.06}  & \tbg{14.38} & \tbg{22.26} & 37.29 &         
                                        \tbb{23.70} &   24.49     &     22.77   & \tbg{14.89} & \tbb{21.86}  \\ 

        \midrule
        \textbf{GaussianFormer-B}     & 30.68       & 19.73       & 19.36       & \tbg{13.19} & \tbb{26.90} & 29.79       & 10.20       & \tbb{15.17} &     12.55  &    9.29      & \tbb{12.96} &       21.45 &     39.55   &                                     23.03 & \tbb{25.07} & \tbb{23.65} & 12.35 & 21.18 \\
        
        \textbf{GaussianFormer-T}     & 31.34       & \tbg{20.42} & \tbg{20.82} & \tbb{12.07} & 26.89       & \tbg{30.94} & \tbb{10.52} & \tbg{16.48} & 13.15 & \tbb{10.46} &     12.90   & \tbb{21.79} & \tbg{41.13} &                                     \tbg{24.22} & \tbg{26.29} & \tbg{24.89}     & 12.80       & 21.45 \\
        
        \textbf{GaussianWorld (ours)} & \tbr{33.40} & \tbr{22.13} & \tbr{21.38} & \tbr{14.12} & \tbg{27.71} & \tbr{31.84} & \tbr{13.66} & \tbr{17.43} & \tbg{13.66} & \tbg{11.46} & \tbr{15.09} & \tbr{23.94} & \tbr{42.98} &                                     \tbr{24.86} & \tbr{28.84} & \tbr{26.74} & \tbr{15.69} & \tbr{24.74} \\
        
        \bottomrule
    \end{tabular}}
    \label{tab:nusc_occ}
    \vspace{-3mm}
\end{table*}

\subsection{3D Gaussian World Model}

We introduce the overall framework of our GaussianWorld.
Starting with 3D Gaussians $\mathbf{g}^{T-1}$ from the previous frame, we initially apply the alignment module $A_{lign}$ to obtain the aligned 3D Gaussians $\mathbf{g}^{T}_{A}$ for the current frame.
In the newly-observed areas, We sample additional 3D Gaussians $\mathbf{g}^{T}_{I}$ and merge them with $\mathbf{g}^{T}_{A}$ to jointly depict the scene: 
\begin{equation}
\begin{aligned}
\label{eq:gaussian-merge}
  \mathbf{g}^{T} = [\mathbf{g}^{T}_{A}, \mathbf{g}^{T}_{I}].
\end{aligned}
\end{equation}
We use the motion layer $M_{ove}$ and the perception layer $P_{er}$ to update $\mathbf{g}^{T}_{A}$ and $\mathbf{g}^{T}_{I}$, respectively, based on the current sensor input $\mathbf{x}^{T}$.
\textit{It is worth noting that the two layers share the same model architecture and parameters, namely the encoder module $E_{nc}$ and the refinement module $R_{ef}$, allowing them to be integrated into a unified evolution layer $E_{vol}$ and computed in parallel.}
This design ensures that GaussianWorld maintains model simplicity and computational efficiency.
We stack $n_{e}$ evolution layers to iteratively refine the 3D Gaussians, endowing the model with sufficient capacity to learn the scene evolution:
\begin{equation}
\begin{aligned}
\label{eq:evolution-layer}
\! \!   \mathbf{g}^{T}_{l+1}\!  =\!  E_{vol}(\mathbf{g}^{T}_{l}, \mathbf{x}_{T}) 
                     \!   = \! \!  \begin{cases}
                     P_{er}(\mathbf{g}^{T}_{l}, \mathbf{x}_{T}), \! \!  &\text{if new,} \\
  M_{ove}(\mathbf{g}^{T}_{l}, \mathbf{x}_{T}),\! \!  &\text{otherwise,}  \\
  
\end{cases}
\end{aligned}
\end{equation}
where $\mathbf{g}^{T}_{l}$ denotes the 3D Guassians of the $l_{th}$ evolution layer, and $l$ ranges from 1 to $n_{e}$.
Furthermore, to address potential misalignments between the 3D Gaussian representation and the real world, we incorporate another $n_{r}$ refinement layers to fine-tune all attributes of 3D Gaussians:
\begin{equation}
\begin{aligned}
\label{eq:refinement-layer}
  \mathbf{g}^{T}_{n+1} &= R_{efine}(\mathbf{g}^{T}_{n}, \mathbf{x}_{T}), \\
                       &= R_{ef}(\mathbf{g}^{T}_{n}; E_{nc}(\mathbf{g}^{T}_{n}, \mathbf{x}_{T}); \{\mathbf{p}, \mathbf{s}, \mathbf{r}, \mathbf{c}, \mathbf{f}\}), \\
\end{aligned}
\end{equation}
where $\mathbf{g}^{T}_{n}$ denotes the 3D Guassians of the $n_{th}$ refinement layer, and $n$ ranges from 1 to $n_{r}$.
The only difference between the evolution layer and the perception layer lies in which attributes of the historical Gaussians are adjusted.
This adjustment can be incorporated into a unified refinement block, as shown in Figure~\ref{fig:network}.
Therefore, both layers can be integrated into a unified Gaussian world layer. 
More details are provided in the supplementary materials.

We adopt the cross entropy loss and the lovasz-softmax~\cite{lovasz} loss for training, and first pretrain our model on the single-frame task.
We then finetune our model with a streaming training strategy, where the images of each scene are input into the model sequentially.
During each iteration of training, the current frame images combined with 3D Gaussians predicted in the last frame are input into the model for 3D occupancy prediction.
The predicted 3D Gaussians in the current frame are passed to the next iteration for continuous streaming training.

In the early stages of streaming training, the model is not yet proficient at predicting the scene evolution, resulting in significant streaming prediction errors.
To enhance training stability, we start training with short sequences and gradually increase the sequence length throughout the training process.
We use probabilistic modeling and randomly discard the 3D Gaussian representation of the previous frame with a probability of $p$ during each iteration.
As training advances, we gradually decrease the value of $p$, allowing the model to adapt to predicting longer sequences.

\section{Experiments}

In this paper, we explore a world-model-based framework to exploit the scene evolution for perception and propose a GaussianWorld model to perform streaming 3D semantic occupancy prediction.
We conducted experiments on the widely used nuScnens~\cite{nuscenes} dataset to evaluate the effectiveness of our GaussianWorld.

\subsection{Datasets}
The nuScenes~\cite{nuscenes} dataset is a public dataset for autonomous driving, which consists of 1000 diverse driving scenes in Boston and Singapore.
These scenes are officially divided into 700, 150, and 150 sequences for training, validation, and testing, respectively.
Each sequence is captured at 20Hz with a 20-second duration, and the keyframes are annotated at 2Hz.
Each sample of a sequence includes multi-view RGB images captured by 6 surrounding cameras.
For the task of 3D semantic occupancy prediction, we utilize the dense 3D semantic occupancy annotations provided by SurroundOcc~\cite{surroundocc} for training and evaluation.
The annotated voxel grids range from -50m to 50m on the X and Y axes and from -5m to 3m on the Z axis, with a resolution of 200 × 200 × 16 for $(H, W, Z)$.
Each grid is assigned one of 18 semantic labels, which include 16 semantic categories, 1 empty category, and 1 unknown category.

\subsection{Evaluation Metrics}
Following common practice~\cite{monoscene}, we use the Intersection over Union (IoU) of all occupied voxels to evaluate the geometry reconstruction performance of the model.
We also report the mean Intersection over Union (mIoU) of all semantic classes to evaluate the semantic perception ability of the model.
The mIoU and IoU can be calculated as follows:
\begin{equation}
\begin{aligned}
\label{eq:miou}
  mIoU &= \frac{1}{|\mathcal{C}|} \sum_{i \in \mathcal{C}} \frac{TP_{i}}{TP_{i} + FP_{i} + FN_{i}}, \\
  IoU &= \frac{TP_{c_0}}{TP_{c_0} + FP_{c_0} + FN_{c_0}},
\end{aligned}
\end{equation}
where $TP_{i}$, $FP_{i}$, $FN_{i}$ are the number of true positive, false positive, and false negative predictions for class $i$, $\mathcal{C}$ is the set of semantic classes, and $c_0$ is the nonempty class.

\subsection{Implementation Details}
Based on previous works~\cite{gaussianformer}, we set the input image resolutions as 900 $\times$ 1600 and employ a ResNet101-DCN~\cite{he2016resnet} initialized from FCOS3D~\cite{fcos3d} as the image backbone.
We utilize a Feature Pyramid Network (FPN)~\cite{FPN} to extract multi-scale image features with downsample sizes of 1/8, 1/16, 1/32, and 1/64.
We utilize a total of 25600 Gaussians to represent 3D scenes and employ 4 Gaussian world layers to refine the attributes of Gaussians, which is consistent with GaussianFormer~\cite{gaussianformer}. 
For optimization, we employ the AdamW~\cite{adamw} optimizer with a learning rate of 4e-4 and weight decay of 0.01.
The model is trained for 20 epochs with a total batch size of 16, consuming 15 hours on 16 NVIDIA RTX 4090 GPUs.

\begin{table*}[t]
    \centering
    \caption{
    \textbf{Comparisons of different temporal modeling methods.}
    The latency and memory consumption for all methods are tested on one NVIDIA 4090 GPU with a batch size of 1 during inference.
    3D Gaussian Fusion and Perspective View Fusion denote the temporal fusion in the 3D Gaussian space and the perspective view space, respectively.
    }
    \vspace{-3mm}
    \setlength{\tabcolsep}{0.02\linewidth}
    \begin{tabular}{c|c|cc|cc}
        \toprule
        Temporal Modeling & Number of Historical Input & Latency & Memory & mIoU & IoU \\
        \midrule
        Single-Frame                & 0 &  \textbf{225} ms & \textbf{6958} M   & 19.73  & 30.68 \\
        3D Gaussian Fusion          & 3 & 379 ms & 9993 M   & 20.24  & 32.27 \\
        Perspective View Fusion     & 3 & 382 ms & 10019 M  & 20.42  & 31.34 \\
        GaussianWorld (ours)        & \textbf{1} & \textbf{228} ms & \textbf{7030} M  & \textbf{21.87}  & \textbf{33.02}  \\
        \bottomrule
    \end{tabular}
    \vspace{-5mm}
    \label{tab:comparisons of temporal models}
\end{table*}

\begin{table}[t!]
    \centering
    \caption{
    \textbf{Ablation on the decomposed factors of GaussianWorld.}
    Neglecting the movements of ego-vehicle or other dynamic objects leads to a performance decline.
    \textbf{×} denotes training collapse without scene completion of newly observed areas.
    }
    \vspace{-3mm}
    \setlength{\tabcolsep}{0.032\linewidth}
    \begin{tabular}{ccc|cc|cc}
        \toprule
        Ego & Dynamics & Completion & mIoU & IoU \\
        \midrule
                   & \checkmark & \checkmark & 18.47  & 28.88 \\
        \checkmark &            & \checkmark & 21.17  & 32.49  \\
        \checkmark & \checkmark &            & \textbf{×}      & \textbf{×}  \\
        \checkmark & \checkmark & \checkmark & \textbf{21.50}  & \textbf{32.72} \\
        \bottomrule
    \end{tabular}
    \vspace{-7mm}
    \label{tab:ablation on the components}
\end{table}

\subsection{Results and Analysis}

\textbf{Vision-Centric 3D Semantic Occupancy Prediction.}
In Table~\ref{tab:nusc_occ}, we present a comprehensive comparison with other state-of-the-art methods for vision-centric 3D semantic occupancy prediction on nuScnene~\cite{nuscenes} validation set, with occupancy labels from SurroundOcc~\cite{surroundocc}.
For the first training stage of GaussianWorld, we replicated GaussianFormer~\cite{gaussianformer} in a single-frame setting, denoted as \textbf{GaussianFormer-B}.
Utilizing just 25600 Gaussians, it achieved comparable performance with current state-of-the-art methods.
In the absence of temporal modeling methods on this benchmark, we introduced a temporal fusion variant of GaussianFormer for a fair comparison, denoted as \textbf{GaussianFormer-T}.
After the second stage of training, our GaussianWorld outperformed all single-frame models and the temporal-fusion-based GaussianFormer by a large margin.
It improved the semantic mIoU by 2.4\% and the geometric IoU by 2.7\% over the single-frame model \textbf{GaussianFormer-B}.
Furthermore, GaussianWorld also outperformed the temporal fusion model \textbf{GaussianFormer-T}, with a 1.7\% increase in mIoU and a 2.0\% increase in IoU.
These results highlight the superiority of our world-model-based framework for perception over the conventional temporal fusion methods.

\textbf{Comparisons of Different Temporal Modeling Methods.}
We provide performance and efficiency comparisons of different temporal modeling methods based on GaussianFormer~\cite{gaussianformer}.
We explored two approaches to implement \textbf{GaussianFormer-T}, performing temporal fusion in the 3D Gaussian space and the perspective view space, respectively.
For the former, we extracted 3D Gaussian representations for each frame independently and utilized 4D sparse convolutions to facilitate temporal interaction between 3D Gaussians across frames.
For the latter, we extracted multi-scale image features for each frame independently and employed deformable attention mechanisms to enable interaction between 3D Gaussians of the current frame and image features from different frames.
As shown in Table~\ref{tab:comparisons of temporal models}, our GaussianWorld surpasses all other temporal modeling methods by a large margin with significantly reduced latency and memory consumption.
Notably, compared to the single-frame model, our GaussianWorld significantly boosts performance with nearly the same inference latency and memory consumption.
This is attributed to our unified and concise model architecture, which can process both single-frame input and streaming inputs without introducing extra computation.

\textbf{Ablation on Decomposed Factors of the Scene Evolution.}
Our GaussianWorld explicitly models the three decomposed factors of the scene evolution for world-model-based perception.
To assess the impact of these factors, we perform ablation studies to confirm their efficacy, as shown in Table~\ref{tab:ablation on the components}.
When not modeling ego motion, our model predicts directly from the previous frame 3D Gaussians without the global affine transformation based on the ego trajectory.
This leads to a 3.0\% decrease in mIoU, highlighting the necessity of ego motion compensation.
Neglecting the movements of dynamic objects impairs our model's ability to represent dynamic scenes, resulting in a slight performance decline.
Without scene completion of newly observed areas, our model relies solely on the initial 3D Gaussians to depict the scene sequence.
As the ego vehicle progresses, all Gaussians eventually fall outside the perception range, leading to training collapse.

\begin{figure}[t]
\centering
\includegraphics[width=0.475\textwidth]{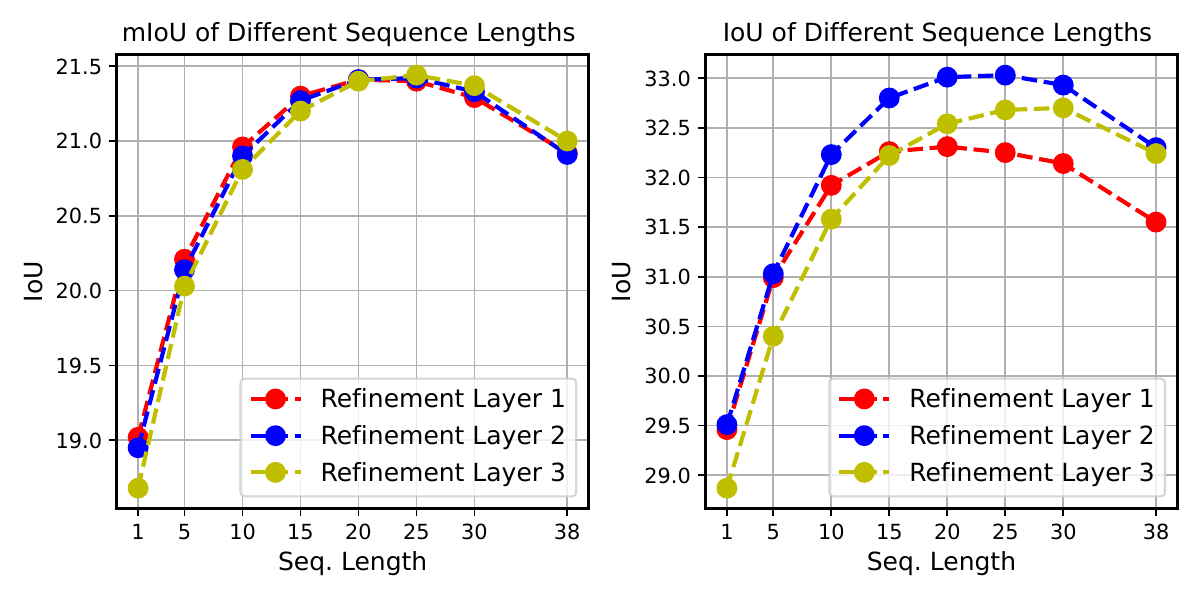}
\vspace{-7mm}
\caption{\textbf{Performance of streaming occupancy prediction with different sequence lengths.}
We also show the performance of using different numbers of refinement layers.
}
\label{fig:iou of different seq lengths}
\vspace{-6mm}
\end{figure}

\textbf{Different Sequence Lengths of Streaming Prediction}
Table~\ref{fig:iou of different seq lengths} shows the mIoU and IoU performance of our GaussianWorld when using different streaming lengths. 
We observe that streaming more frames generally leads to better performance yet slightly drops after around 20 frames.
The improvement results from our modeling of scene evolution to effectively consider history frames.
We think the performance drop is due to the quality of the annotations.
Existing 3D occupancy annotations are collected by fusing multi-frame LiDAR, leading to sparsity at the edge of the scene.

We also observe a trade-off in using different numbers of refinement layers.
This is because using more refinement layers focuses more on the dynamics modeling but will damage the prediction consistency of static elements.

\begin{table*}[t]
    \centering
    \caption{
    \textbf{Ablation on the schedules of streaming training.}
    Seq. Lengths and Epochs denote the increasing sequence length and the corresponding number of epochs during training.
    Sum and Mean represent gradient accumulation by summation and averaging, and None indicates no gradient accumulation.
    Prob. denotes whether to use probabilistic modeling.
    }
    \vspace{-3mm}
    \setlength{\tabcolsep}{0.02\linewidth}
    \resizebox{1.0\linewidth}{!}{
    \begin{tabular}{c|cccc|c|cc}
        \toprule
        Index & Seq. Lengths & Epochs & Grad. Acc. & Prob. & Training Time & mIoU & IoU \\
        \midrule
        A  & [5, 10, 20, 30, 38]  & [80, 40, 20, 20, 20] & Sum   & N & 30 h  & 19.77  & 30.90 \\
        B  & [5, 10, 20, 30, 38]  & [80, 40, 20, 20, 20] & Mean  & N & 30 h  & 19.81  & 30.93 \\
        C  & [5, 10, 20, 30, 38]  & [40, 20, 10, \ \ 5, 20] & None & N & 20 h  & 18.63  & 28.69 \\
        D  & [5, 10, 20, 30, 38]  & [\ \ 0,\ \ \  0,\ \ \ 0,\ \ \ 0, 30] & None & N & \textbf{15} h  & 19.82  & 30.96  \\
        E  & [5, 10, 20, 30, 38]  & [\ \ 5,\ \ \  5,\ \ \ 5,\ \ \ 5, 20] & None & Y & \textbf{15} h  & \textbf{20.24}  & \textbf{31.22}  \\
        \bottomrule
    \end{tabular}}
    \vspace{-3mm}
    \label{tab:training strategy}
\end{table*}

\begin{figure*}[t]
\centering
\includegraphics[width=\textwidth]{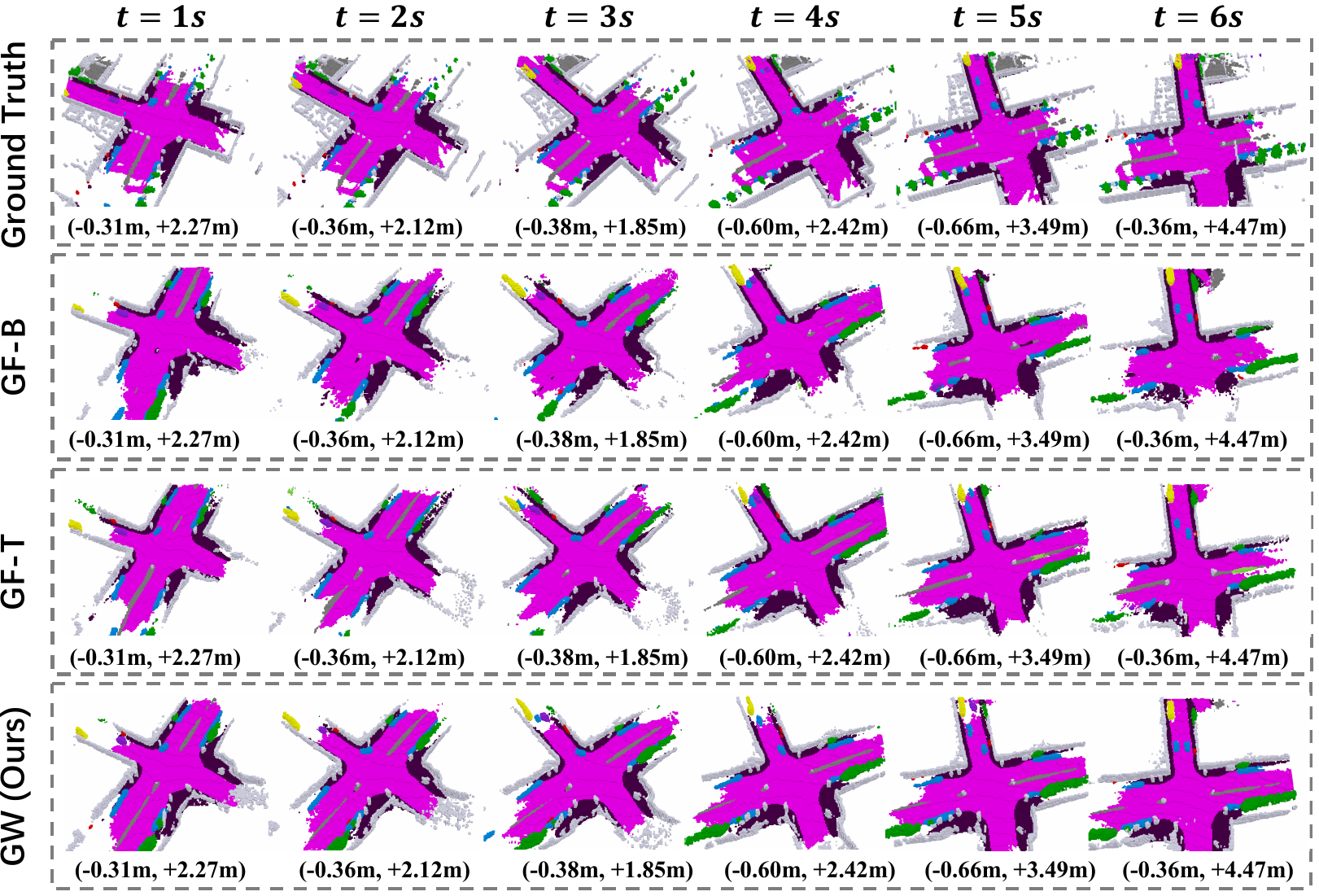}
\vspace{-7mm}
\caption{\textbf{Visualizations of the results of different temporal modeling methods.}
GF-B, GF-T, and GW denote GaussianFormer-B, GaussianFormer-T, and GaussianWorld, respectively.
The numbers under each image indicate the ego trajectory of each frame.
}
\label{fig:vis}
\vspace{-6mm}
\end{figure*}

\textbf{Ablation on the Schedules of Streaming Training}
In Table~\ref{tab:training strategy}, we explore the impact of different streaming training strategies.
Due to the poor performance of training directly on long sequences, we gradually increase the sequence length during training.
Although accumulating gradients across different frames within a sequence can lead to better optimization, it results in low training efficiency. 
Probabilistic modeling of sequence lengths allows the model to handle sequences of varying lengths and gradually adapt to longer sequence lengths, effectively enhancing the streaming prediction performance.

\textbf{Visualization Results.}
We provide a qualitative analysis of our GaussianWorld compared with existing methods.
We observe that the predictions of our methods show more cross-frame consistency than the baselines, especially for the static elements.
This verifies the effectiveness of our Gaussian-based explicit streaming modeling.

\section{Conclusion}
In this paper, we have presented a world-model-based framework to exploit the scene evolution for 3D semantic occupancy prediction. 
We reformulate 3D occupancy prediction as a 4D occupancy forecasting problem conditioned on the current sensor input.
We decompose the scene evolution into three factors and exploit the explicity of 3D Gaussians to model them effectively and efficiently.
We then employ a Gaussian world model (GaussianWorld) to explicitly exploit the scene evolution in the 3D Gaussian space and facilitate 3D semantic occupancy prediction in a streaming manner.
Our model demonstrates state-of-the-art performance compared to existing methods without introducing additional computation overhead.
It is an interesting future direction to apply our model to other perception tasks.

\textbf{Limitations.}
Our model cannot achieve full cross-frame consistency for static scenes, resulting from the inaccurate disentanglement between dynamic and static elements.

\appendix

\twocolumn[{%
\renewcommand\twocolumn[1][]{#1}%
\vspace{-2mm}
\vspace{-2mm}
\begin{center}
    \centering
    \includegraphics[width=\linewidth]{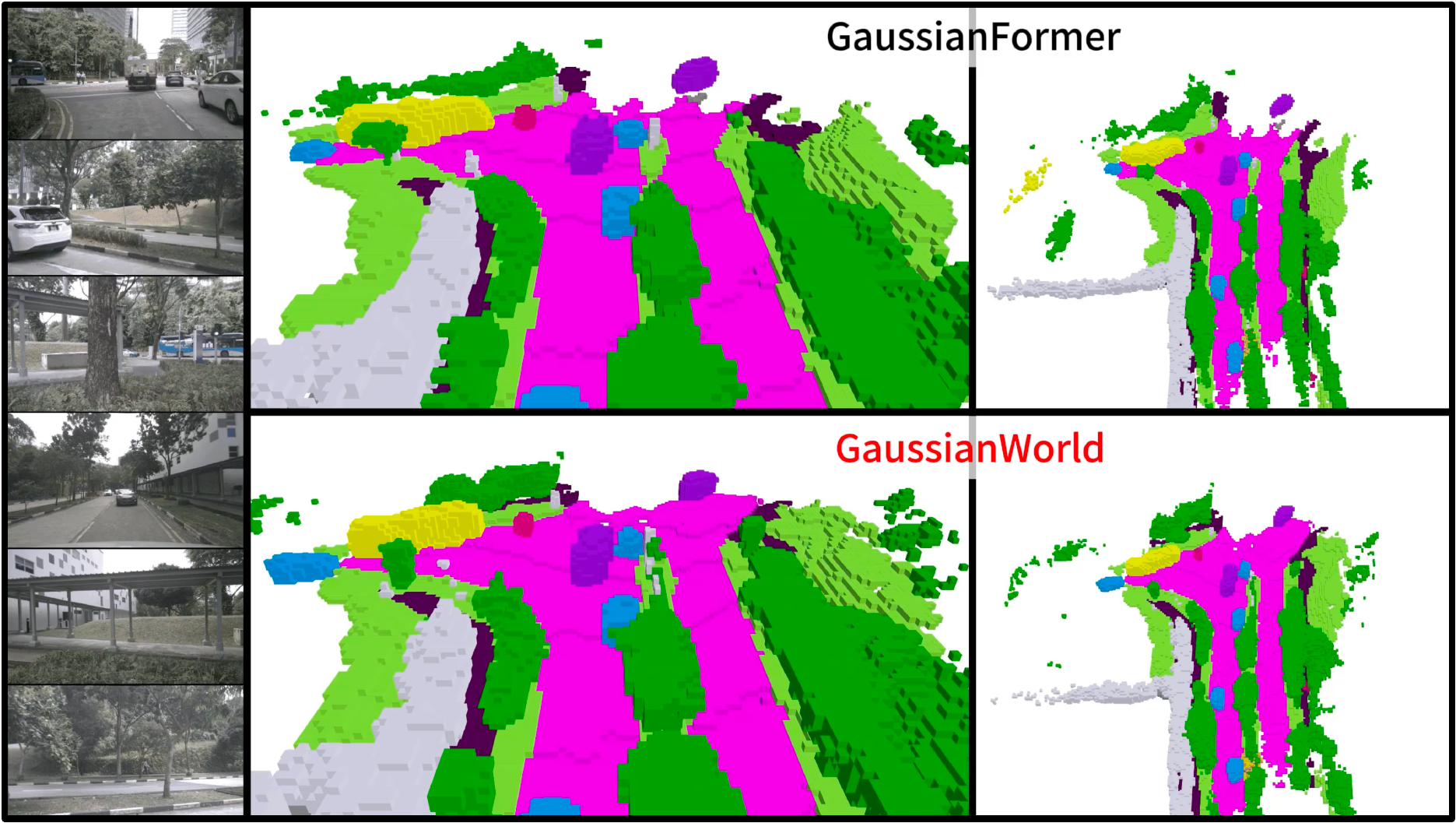}
    \vspace{-7mm}
    \captionof{figure}{
\textbf{Visualizations of the proposed GaussianWorld compared to GaussianFormer~\cite{gaussianformer} for 3D semantic occupancy prediction on the nuScenes~\cite{nuscenes} validation set.}
We visualize the six surrounding camera inputs and the corresponding occupancy prediction results.
The upper row shows the predicted occupancy by GaussianFormer in the global view(left) and the bird's eye view(right).
The lower row shows the prediction results of GaussianWorld.
}
\label{video}
\end{center}%
}]

\section{Additional Implementation Details}
\textbf{Evolution Layer.}
We employ a unified evolution layer to model the evolution of aligned historical Gaussians and the perception of newly-completed Gaussians.
As shown in Figure~\ref{fig:framework}, the evolution layer consists of a self-encoding module, a cross-attention module, and a unified refinement block.
Specifically, we first voxelize all 3D Gaussians and utilize a 3D sparse convolution block to facilitate interaction between 3D Gaussians. 
We then adopt deformable attention to enable interaction between 3D Gaussians and multi-scale image features. 
Finally, we use a unified refinement block to update attributes of historical Gaussians and new Gaussians separately.
To refine only the positions of dynamic historical Gaussians, we use the semantic probability of Gaussians as a soft semantic weight to update their positions.

\textbf{Unified Refinement Block.}
In this module, a unified prediction layer is first employed to predict the attribute modifications for all Gaussians.
For newly-completed Gaussians, the predicted modifications are directly incorporated into their original attributes.
For historical Gaussians, only the positions of the dynamic Gaussians are updated.
This is accomplished by using the probability of the Gaussian's dynamic semantic category as semantic weights to update the position of these Gaussians, which corresponds to the evolution mode of the unified refinement block.

\textbf{Refinement Layer.}
To address the misalignment between the 3D Gaussian representation and the real world, we also employ a unified refinement layer to fine-tune all attributes of all 3D Gaussians. 
The only difference from the evolution layer is that we use an additional temporal weight to update all attributes of historical Gaussians, where the unified refinement block switches to the perception mode, as shown in Figure~\ref{fig:network}.

\section{Additional Experiments}
We have refined the model architecture of Gaussianformer~\cite{gaussianformer} with several key modifications.
To learn the scene evolution, we introduce an additional temporal feature attribute to capture the historical information of 3D Gaussians.
Rather than predicting the updated properties directly, we predict the changes to Gaussian properties, thereby preserving their original characteristics as much as possible while modeling the scene evolution.
Considering that Gaussians need a broad range of movement to model dynamic object motion, we expand the influence range of Gaussians when interacting with images and predicting occupancy.
We conduct ablation studies to validate the effectiveness of these designs.
As shown in Table~\ref{tab:ablation on the model structure}, the absence of these designs results in a slight performance degradation.

\begin{table}[h]
    \centering
    \caption{
    \textbf{Ablation on the model structure design.}
    Temp. Feat., Delta Ref., and Range denote utilizing the additional temporal feature property, predicting the changes to Gaussian properties, and expanding the influence range of Gaussians, respectively.}
    \vspace{-3mm}
    \setlength{\tabcolsep}{0.032\linewidth}
    \begin{tabular}{ccc|cc}
        \toprule
        Temp. Feat. & Delta Ref. & Range & mIoU & IoU \\
        \midrule
                   & \checkmark & \checkmark & 21.55  & 32.81 \\
        \checkmark &            & \checkmark & 21.37  & 32.25  \\
        \checkmark & \checkmark &            & 21.21  & 32.32  \\
        \checkmark & \checkmark & \checkmark & \textbf{21.87}  & \textbf{33.02} \\
        \bottomrule
    \end{tabular}
    \vspace{-3mm}
    \label{tab:ablation on the model structure}
\end{table}

\section{Video Demonstration}
Figure~\ref{video} shows a sampled image from the video demo for 3D semantic occupancy prediction on the nuScenes~\cite{nuscenes} validation set.
Compared to GaussianFormer~\cite{gaussianformer}, our GaussianWorld shows more cross-frame consistency, especially for static elements.
This demonstrates the effectiveness of our Gaussian-based explicit streaming modeling.

\end{document}